\documentclass[letterpaper, preprint, paper,11pt]{AAS}	% for preprint proceedings

\usepackage{bm}
\usepackage{amsmath}
\usepackage{amssymb}
\usepackage{subfigure}
\usepackage{float}
\usepackage[colorlinks=true, pdfstartview=FitV, linkcolor=black, citecolor= black, urlcolor= black]{hyperref}
\usepackage{overcite}
\usepackage{footnpag}			      	% make footnote symbols restart on each page
\usepackage{enumitem}

\usepackage{soul}
\PaperNumber{25-553}

\begin{document}

%\title{Dynamics and Control of a Dual-Arm Space Manipulator System for Autonomous Robot-to-Robot Handover in
%Space Operations}

% \title{A Modeling and Baseline Control Framework for Autonomous Space Manipulator Systems in Robotic Handover Operations}

\title{Modeling and Control Framework for Autonomous Space Manipulator Handover Operations}

\author{Diego Quevedo\thanks{PhD Student, Aerospace Engineering and Engineering Mechanics, University of Cincinnati, 2850 Campus Way, Cincinnati, OH 45221-0070.}, Sarah Hudson\thanks{PhD Student, Aerospace Engineering and Engineering Mechanics, University of Cincinnati, 2850 Campus Way, Cincinnati, OH 45221-0070.},
\ and Donghoon Kim\thanks{Assistant Professor,Aerospace Engineering and Engineering Mechanics, University of Cincinnati, 2850 Campus Way, Cincinnati, OH 45221-0070.}
}

\maketitle{} 		

%\hl{comment: see the revised title and let me know what you think.}
%\Diego{Is it "a modeling framework" and "a baseline control framework"? Framework to me just means that it's early work with low scope/goal, which would be more honest. Is that what you're going for?}\Kim{Let's finalize the title after completing the paper.}

%%%%%%%%%%%%%%%%%%%%%%%%%%%%%%%%%%%%%%%%%%%%%%%%%%%%%%%%%%%%%%%%%%%%%%%%%%%%%%%%
\begin{abstract}
Autonomous space robotics is poised to play a vital role in future space missions, particularly for In-space Servicing, Assembly, and Manufacturing (ISAM). A key capability in such missions is the Robot-to-Robot (R2R) handover of mission-critical objects. This work presents a dynamic model of a dual-arm space manipulator system and compares various tracking control laws. The key contributions of this work are the development of a cooperative manipulator dynamic model and the comparative analysis of control laws to support autonomous R2R handovers in ISAM scenarios.
\end{abstract}
%%%%%%%%%%%%%%%%%%%%%%%%%%%%%%%%%%%%%%%%%%%%%%%%%%%%%%%%%%%%%%%%%%%%%%%%%%%%%%%%
\section{Introduction}
The global space industry has grown significantly over the past decade and is expected to continue expanding \cite{wef}. In-space Servicing, Assembly, and Manufacturing (ISAM) is emerging as a transformative approach to space operations \cite{nasaIsam}. Rather than deploying a large, monolithic spacecraft, future missions could involve launching modular components for %on-orbit 
assembly or servicing, such as refueling or repairs. This approach could have %fundamentally 
changed the deployment of complex spacecraft like the James Webb Space Telescope, which instead could have been incrementally assembled in orbit \cite{aerospaceIsam}.

A key enabler of ISAM is the advancement of autonomous space robotics \cite{boyu}. Robotic systems may be tasked with inspection, capture,  refueling, repair, or assembly operations. Incorporating multiple robotic arms enhances flexibility, particularly for dual-arm tasks like robotic handover. In these scenarios, one arm may pass an object to the other, expanding the system's effective workspace. This makes Robot-to-Robot (R2R) handover a critical capability. Despite its importance, existing literature often lacks rigorous analysis in space context, focusing instead on ground-based systems or human-robot interactions \cite{WU2023,Ortenzi2021,Colombo2024,Sileo2021}.

Dynamic modeling for Space Manipulator Systems (SMS) is well-established, often using energy-based methods such as Lagrange dynamics \cite{RODRIGUES2021, JIA2019, WILDE2018}. However, high Degree-Of-Freedom (DOF) systems pose computational challenges, which can be addressed through symbolic computation tools or by employing numerical methods, such as recursive Newton-Euler techniques.

A key consideration in dynamic modeling is whether the system is free-flying or free-floating. The former involves a controlled base spacecraft, while the latter assumes the base is uncontrolled \cite{PAPADOPOULOS2021}. Free-floating systems are typically modeled by enforcing conservation of %total 
linear and angular momentum. This implies that the system's center of mass remains fixed in inertial space, and %that 
all body motions are coupled-movement of one manipulator induces reactive motion in the rest of the system.

This dynamic coupling can be leveraged by designing one manipulator as a task arm the other as a balancer arm. Additionally, free-floating systems exhibit nonholonomic behavior, meaning their final configuration (i.e., position and orientation) depends on the path taken, not just the endpoint. Thus, an end-effector moving to a fixed point in inertial space may cause the base to end in different orientations depending on the motion path. In contrast, free-flying systems with a stabilized base simplify modeling and control, resembling ground-based manipulators.

A control law calculates the torque input needed to achieve a desired joint trajectory. Control techniques for SMS vary based on system goals, manipulator redundancy, and whether the system is free-flying or free-floating. Nanos and Papadopoulos \cite{NANOS2017} proposed a %P
{p}roportional-%D
{d}erivative %(PD) 
controller capable of compensating for non-zero angular momentum in free-floating SMS during point-to-point tracking. A Nonlinear Model Predictive Control {(NMPC)} law incorporate{s} the SMS model and optimizes a cost function over a future prediction horizon to calculate control inputs \cite{guo, RYBUS2016, SHI2017}. This approach allows NMPC to handle system constraints and nonlinear dynamics while minimizing objectives such as energy usage and end-effector pose error.

%These functions typically aim to minimize joint tracking but may include criteria such as control effort reduction.
%\Diego{A desired joint trajectory should first be generated to send the joint error to the NMPC controller.}.
%\hl{comment: How can we determine the desired tracking trajectory of joints for the free-floating system in real-time? I’m not sure this can be used as an objective function. If my understanding is incorrect, could you please clarify the reasoning?} 
%\hl{Still confusing... I can't understand what you would like to address here.}

To support R2R handover for ISAM, this work develops a dual-arm robotic model and evaluates its performance in a mock handover scenario. The robot tracks online-computed end-effector trajectories using NMPC to evaluate its effectiveness, and the results are compared against a baseline Proportional-Integral-Derivative (PID) controller to assess NMPC's effectiveness. Although this work focuses on NMPC, the findings may serve as a baseline for future artificial intelligence and machine learning-based control research. The goal is to explore both benefits and limitations of NMPC in this context.

%%%%%%%%%%%%%%%%%%%%%%%%%%%%%%%%%%%%%%%%%%%%%%%%%%%%%%%%%%%%%%%%%%%%%%%%%%%%%%%%%%%%%%%%%%%%%%%%%%%%%%%%%%%
\section{Problem Statement}
The goal of the SMS in this work is to perform an object handover between the manipulators, while allowing the free-floating base to dynamically react. A dynamic model of a complete SMS must be developed. The system operates in microgravity as a free-floating platform, meaning the base spacecraft is not rigidly attached to an inertial frame. Consequently, any internal motion induces a coupled reaction. These dynamic couplings must be accounted for in the SMS trajectory planning by calculating Inverse Kinematics (IK).

%The big goal is to conduct R2R handover, to achieve this we need to track end-effector pose, and to achieve that we conduct inverse kinematics to solve for a configuration that is then tracked.
To achieve precise control, the SMS end-effectors must be controllable in joint space. This requires defining a trajectory of poses for the end-effectors to follow
and computing the corresponding control inputs. Due to the system’s nonlinear and coupled dynamics, nonlinear controllers are necessary. Both manipulators are modeled after the Kinova Gen3 7-DOF manipulator. 

The following assumptions are made:
\begin{itemize}
[topsep=0pt,itemsep=-1ex,partopsep=1ex,parsep=1.2ex]
    \item The SMS is in a geostationary orbit around Earth.
    \item Disturbances like gravity gradient torque, J2 perturbations, and solar radiation pressure are negligible and thus ignored, as the simulation is much shorter than the SMS orbital period.
    \item The SMS has perfect knowledge of the deputy's pose.
    \item The deputy is assumed massless to minimize dynamic coupling during handover.
\end{itemize}

%%%%%%%%%%%%%%%%%%%%%%%%%%%%%%%%%%%%%%%%%%%%%%%%%%%%%%%%%%%%%%%%%%%%%%%%%%%%%%%%%%%%%%%%%%%%%%%%%%%%%%%%%%%
\section{Methodology}
The overall SMS is a dual-arm system, with each manipulator having 7-DOF. The Kinova Gen3 7-DOF robot arm is used as the manipulator in this work. Each manipulator is redundant, meaning it has more DOF than necessary to attain an arbitrary position in 3-Dimensional (3D) space.

%%%%%%%%%%%%%%%%%%%%%%%%%%%%%%%%%%%%%%%%%%%%%%%%%%%%%%%%%%%%%%%%%%%%%%%%%%%%%%%%%%%%%%%%%%%%%%%%%%%%%%%%%%%
\subsection{Kinematics}
\begin{figure}[!t]
    \centering
    \includegraphics[width=0.75\linewidth]{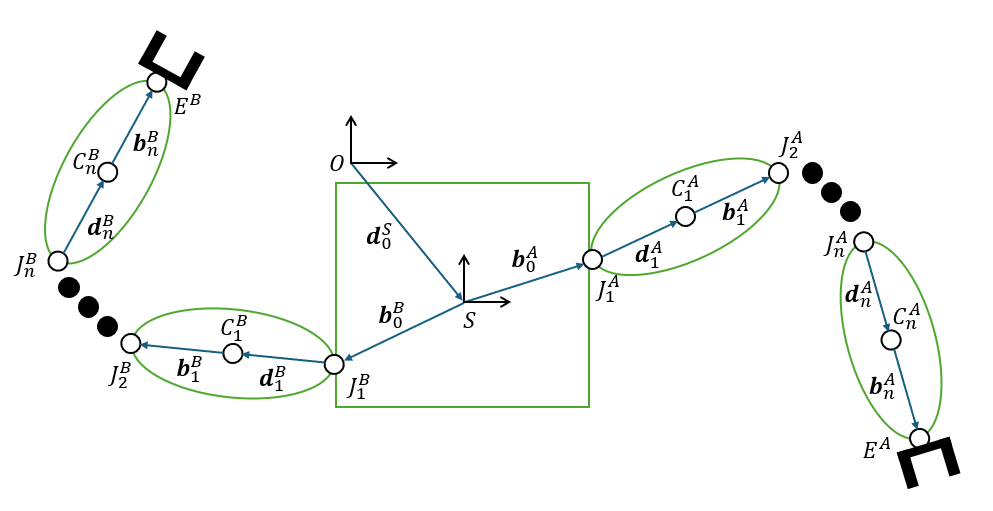}
    \vspace{-5mm}
    \caption{Kinematic Diagram of 20-DOF SMS}
    \label{fig:kin-dia}
\end{figure}

To determine the inertial position of each end-effector and Center of Mass (CoM) of each rigid body, the system's kinematics are computed. Figure \ref{fig:kin-dia} illustrates how each body's inertial position is derived. Note that the system is 3D, but the diagram simplifies the frame depictions for clarity.

The inertial frame is denoted as $O$, the base spacecraft and its attached body frame as $S$, and the CoM of each manipulator link $C_i^k$, where $k:(A,B)$ and $i:[1,n]$. The frames associated with the joints and end-effectors are defined as $J_i^k$ and $E^k$, respectively.

Position vectors $\textbf{d}_j^p$ and $\textbf{b}_j^p$, where $p:(S,A,B)$ and $j:[0,n]$, connect the frames. Typically, $\textbf{d}_j^p$ go from a joint to the COM of the corresponding link, while $\textbf{b}_j^p$ span from a link's COM to the next joint. These vectors represent the relative positions from COM to joint and vice versa. The end-effectors are assumed to be rigidly attached to their respective links at $E^k$.

The positions of $S$, $C_i^k$, and $J_i^k$ must be calculated relative to the inertial frame. Also, $\textbf{d}_j^p$ and $\textbf{b}_j^p$ are easily defined in their respective frames, i.e., $\textbf{d}_1^B$ is most conveniently expressed in $J_1^B$. Therefore, it is necessary to transform the relative vectors to a common frame %in order 
to perform vector addition. This is accomplished using a transformation matrix, like a direction cosine matrix%, to shift frames
.

The inertial position of every rigid body COM can be described with the following:
\begin{equation}
    \textbf{r}_{i/N}^k = \textbf{d}_0^S + \sum_{j=1}^{i} ( \textbf{b}_{j-1}^k +  \textbf{d}_j^k)
\end{equation}

In order to calculate the forward kinematics of the SMS, it is also necessary to know the inertial positions of each joint. The inertial position of each joint is calculated as
\begin{equation}
    \textbf{p}_{i/N}^k = \textbf{d}_0^S + \sum_{j=1}^{i} (\textbf{b}_{j-1}^k + \textbf{d}_{j-1}^k)
\end{equation}

The inertial position of the end-effector position is given by
\begin{equation}\label{eq:r_en2}
    \textbf{r}_{E/N}^k = \textbf{d}_0^S + \sum_{j=1}^{n} ( \textbf{b}_{j-1}^k +  \textbf{d}_j^k) + \textbf{b}_n^k
\end{equation} 
The second term yields the position vector to the COM of the last link; the third term is added to reach the end-effector joint. Although not required, an additional position vector can be added from this joint to obtain the inertial location of the tool, which may lie between the gripper fingers. 

To derive the %inverse kinematics
IK of the end-effector, the time derivative of Eq.~\eqref{eq:r_en2} must be taken as follows:
\begin{equation}
    \textbf{v}_{E/N}^k = \frac{^Nd}{dt}(\textbf{r}_{E/N}^k)
\end{equation}

The forward kinematics of each end-effector can then be described as:
\begin{equation}
    \textbf{v}_{E/N}^k = J^k\dot{\textbf{q}}^k
\end{equation}
where $J^k\in \mathbb{R}^{6\times (6+n)}$ is the geometric Jacobian matrix. The general state vector $\textbf{q}^k \in \mathbb{R}^{(6+n) \times 1}$ for the $k$-th arm is defined as:
\begin{equation}
    \textbf{q}^k=
    \begin{bmatrix}
        \textbf{d}_0^S & \boldsymbol{\theta}^S & \textbf{q}_m^k
    \end{bmatrix}^T
\end{equation}
where
\begin{equation}\label{eq:base_pos}
    \textbf{d}_0^S =
    \begin{bmatrix}
        q_1^S & q_2^S & q_3^S
    \end{bmatrix}
\end{equation}
\begin{equation}\label{eq:base_ang}
    \boldsymbol{\theta}^S =
    \begin{bmatrix}
        q_4^S & q_5^S & q_6^S
    \end{bmatrix}
\end{equation}
\begin{equation}\label{eq:manip_state}
    \textbf{q}_m^k =
    \begin{bmatrix}
        q_1^k & q_2^k & q_3^k & q_4^k & q_5^k & q_6^k & q_7^k
    \end{bmatrix}^T
\end{equation}
Here, Eq.~\eqref{eq:base_pos} refers to the inertial position of the base spacecraft, Eq.~\eqref{eq:base_ang} the base spacecraft's body-fixed frame Euler angles relative to the inertial frame, and Eq.~\eqref{eq:manip_state} the manipulator joint states. Each generalized coordinate can be combined into a single vector, $\textbf{q} \in \mathbb{R}^{(6+2n) \times 1}$ for later use:
\begin{equation}
    \textbf{q} = 
    \begin{bmatrix}
        \textbf{d}_0^S & \boldsymbol{\theta}^S & \textbf{q}^A & \textbf{q}^B
    \end{bmatrix}
\end{equation}

The IK of an end-effector requires inverting the geometric Jacobian. Since $J^k$ is not square in this case, the Moore-Penrose pseudoinverse, denoted by $\dagger$, is used:
\begin{equation}
    \dot{\textbf{q}}^k = J^{k^{\dagger}}\textbf{v}_{E/N}^k
\end{equation}

%%%%%%%%%%%%%%%%%%%%%%%%%%%%%%%%%%%%%%%%%%%%%%%%%%%%%%%%%%%%%%%%%%%%%%%%%%%%%%%%%%%%%%%%%%%%%%%%%%%%%%%%%%%
\subsection{Dynamics}
A classical approach to deriving the dynamics of an SMS is to use the Lagrangian method, which is based on the principle of virtual work\cite{WILDE2018}. The Lagrangian of a system is defined as the sum of kinetic energy minus the sum of potential energy. Since the SMS operates in microgravity, the potential energy of the system can be neglected. Therefore, the Lagrangian is the sum of each body's linear and rotational kinetic energy.

\begin{equation}\label{eq:lagrangian}
    L = T = \frac{1}{2}\sum_{i=1}^n\sum_{k=A}^B((\boldsymbol{\omega}_i^k)^T \textbf{I}_i^k \boldsymbol{\omega}_i^k + m_i (\dot{\textbf{r}}^k)^T \dot{\textbf{r}}^k)
\end{equation}

To derive the dynamics, the Lagrangian above must be manipulated as follows:
\begin{equation}\label{eq:lagrange_method}
    \frac{d}{dt}\frac{\partial L}{\partial \dot{\textbf{q}}} - \frac{\partial L}{\partial \textbf{q}} = \boldsymbol{Q}
\end{equation}
where the generalized coordinate vectors $\textbf{q}$ and $\dot{\textbf{q}}$ are the complete generalized coordinates and are therefore $[(6+2n) \times 1]$ vectors. Similarly, the vector $\textbf{Q}$ is a $[(6+2n) \times 1]$ generalized input vector for each joint. The first six inputs, $[Q_1 \ldots Q_6]$, are equal to zero because the SMS is assumed to be free-floating and thus cannot exert input forces or torques. Eq.~\eqref{eq:lagrange_method} is solved symbolically using the generalized coordinates and their time derivatives, resulting in a $[(6+2n) \times 1]$ system of equations.

From the system of equations derived via Eq.~\eqref{eq:lagrange_method}, each coordinate  can be isolated and rearranged to yield the following, known as the general robotic manipulator equation of motion {for SMS}:
\begin{equation} \label{eom}
    H(\textbf{q})\ddot{\textbf{q}} + C(\textbf{q},\dot{\textbf{q}})\dot{\textbf{q}} = \textbf{Q}
\end{equation}
where $H(\textbf{q})$ and $C(\textbf{q},\dot{\textbf{q}}$) are $[(6+2n)\times (6+2n)]$ matrices, referred to as the inertia matrix and Coriolis matrix, respectively. %Also, $\textbf{G}(\textbf{q})$ is the $(6+2n)$-sized gravity vector (which is neglected going forward), and 
$\textbf{Q}$ is the $(6+2n)$-sized general input vector to the system. The vectors $\textbf{q}$, $\dot{\textbf{q}}$, and $\ddot{\textbf{q}}$ are the $(6+2n)$-sized position, velocity, and acceleration states of the system, respectively. These state vectors contain the DOF information for every joint, from the base to both end-effectors. Literature often further breaks down the $H$ matrix into coupling components \cite{WANG2018}:
\begin{equation}
    \begin{bmatrix}
        H_b & H_{bm}^A & H_{bm}^B \\
        H_{bm}^A & H_m^A & \bm0_n \\
        H_{bm}^B & \bm0_n & H_m^B
    \end{bmatrix}
    \ddot{\textbf{q}}
    +
    \begin{bmatrix}
        C_b \\
        C_m^A \\
        C_m^B
    \end{bmatrix}
    \dot{\textbf{q}}
    =
    \begin{bmatrix}
        \bm 0_{6\times 1} \\
        \boldsymbol{\tau}^A \\
        \boldsymbol{\tau}^B
    \end{bmatrix}
\end{equation}
where $H_b$ is the $[6\times6]$ inertia matrix of the base spacecraft, $H^k_m$ are the $[n\times n]$ inertia matrices of the manipulators, $H^k_{bm}$ are the $[6\times n]$ inertia coupling matrices between the base spacecraft and the manipulators. Also, $C_b$ is the $[6\times20]$ Coriolis matrix of the base, $C_m^k$ are the $[7\times20]$ Coriolis matrices of the manipulators, and $\boldsymbol{\tau}^k$ are $[7\times1]$ are the torque vectors.

To use Eq.~\eqref{eom} to solve for the system's motion one must compute the generalized coordinate acceleration. This acceleration can then be integrated twice to find the velocity and position over time:
\begin{equation} \label{ode}
    \ddot{\textbf{q}} = H(\textbf{q})^{-1}[\textbf{Q} - C(\textbf{q},\dot{\textbf{q}})\dot{\textbf{q}}]
\end{equation}

Using Eq.~\eqref{ode}, the input $\textbf{Q}$ is selected under the assumption that the initial state positions and velocities are known, allowing the calculation of the acceleration. Integration of Eq.~\eqref{ode} over time yields the system's motion. However, this method is computationally expensive , as inverting the inertia matrix is costly, especially for high DOF systems, such as the one considered in this work. As a result, a numerical approach is preferred over the classical analytical approach for dynamic calculations.

\begin{figure}[!t] 
    \centering
    \includegraphics[width=1\linewidth]{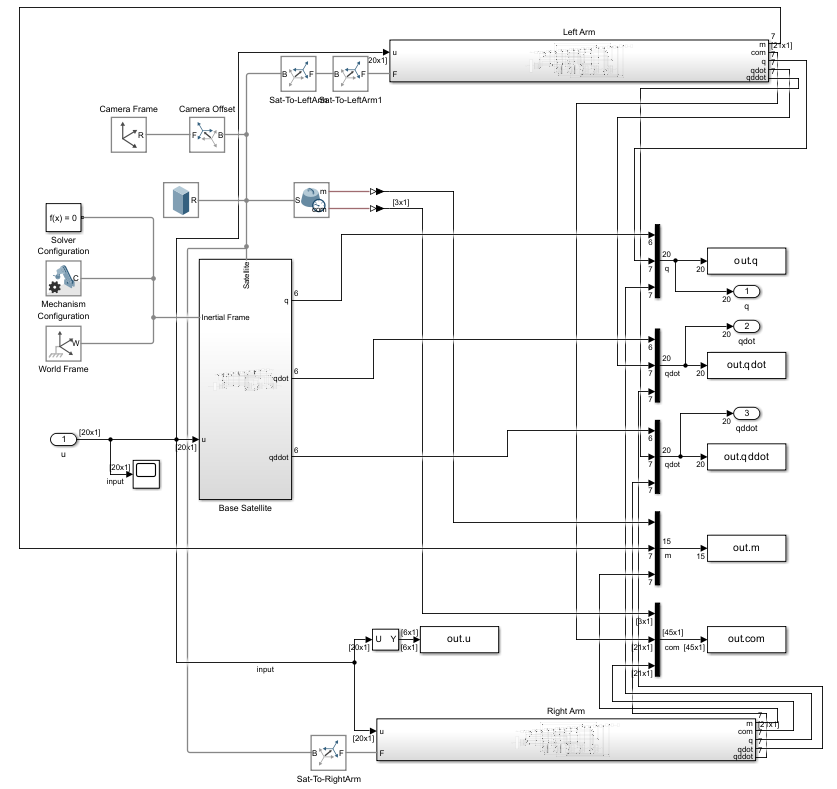}
    \caption{20-DOF SMS Simulink Block Diagram}
    \label{fig:model}
\end{figure}

MATLAB/Simulink provides a more computationally efficient way to calculate robot dynamics using \emph{Multi-Body Simscape}. In this tool, the rigid bodies of the system are represented visually and connected via wires. Blocks representing the robot's joints are connected between transforms, allowing the joint poses to be described. A key advantage of using Simscape is its visualizer, which enables simulation of robot's motion. Figure \ref{fig:model} illustrates the high-level model of the 20-DOF SMS within Simulink.
%%%%%%%%%%%%%%%%%%%%%%%%%%%%%%%%%%%%%%%%%%%%%%%%%%%%%%%%%%%%%%%%%%%%%%%%%%%%%%%%%%%%%%%%%%%%%%%%%%%%%%%%%%%
\subsection{Controller}
The PID controller is a classical feedback control law that provides a control signal to a system plant to reduce a variable's error \cite{MANDAVA2015, SLOTINE1991}. PID controllers are popular in practical applications due to their simplicity. The PID controller used in this work, $\textbf{u} \in \mathbb{R}^{2n}$, where $2n$ is the total DOF of the dual manipulators, is defined as:
\begin{equation}\label{eq:PID}
    \textbf{u} = K_P\textbf{e} + K_I\int\textbf{e}+K_D\dot{\textbf{e}}
\end{equation} 
where $\textbf{e}\in \mathbb{R}^{2n}$ is the combined joint errors of both 7-DOF manipulators, given by $\textbf{e} = \textbf{q} - \textbf{q}_d$. Here, the subscript d indicates the desired state. Also, $K_P \in \mathbb{R}^{2n\times2n}$ is the proportional gain matrix, $K_I \in \mathbb{R}^{2n\times2n}$ is the integral gain matrix, and $K_D \in \mathbb{R}^{14\times14}$ is the derivative gain matrix, with each matrix being diagonal. To simplify tuning, all diagonal entries can be set to the same value, reducing the problem to tuning just three scalar gains. More sophisticated methods may be used to find optimal gains and adapt them to working conditions on a joint-by-joint basis \cite{NGUYET2023}.

Controllers like PID work best for linear systems, which real-world systems, like the SMS, are not. Linear control laws rely on the assumption that the operating range of the controlled variables remains small, especially when applied to nonlinear systems \cite{SLOTINE1991}. Therefore, using PID control for the SMS does not guarantee stability, and the accuracy of the controller may be  degraded compared to that of nonlinear controllers

A feedback NMPC is suitable for end-effector trajectory tracking of an SMS. It has been shown that NMPC outperforms other nonlinear controllers for dual-arm SMS in the absence of disturbances \cite{SHI2017}. NMPC is an optimal controller that minimizes trajectory tracking error by using the following cost function from the literature\cite{guo}:
\begin{equation} \label{eq:cost}
    J_\text{cost}=\frac{1}{2}\int_{T_1}^{T_2}\textbf{e}(t+\tau)^T\textbf{e}(t+\tau)d\tau
\end{equation}
and the resulting control law derived from Eq.~\eqref{eq:cost} is:
\begin{equation} \label{eq:control}
    \textbf{u}_\text{m} = -H(\textbf{q})[A_1\textbf{e}(t)+A_2\dot{\textbf{e}}(t)]+C(\textbf{q},\dot{\textbf{q}})\dot{\textbf{q}}+H(\textbf{q})\ddot{\textbf{q}}(t)%.
\end{equation}
The vector $\textbf{u}_\text{m} \in \mathbb{R}^{2n}$ corresponds to the input $\textbf{Q}$ in Eq.~\eqref{eom}. The constant matrices $A_1$ and $A_2$ are gains on the error, set as $A_1 = \frac{10}{3T_r^2}I_{14\times14}$ and $A_2 = \frac{5}{2T_r}I_{14\times14}$, where $I$ is the identity matrix and $T_r$ is the rolling period. The rolling period acts as the time horizon for the NMPC and functions like a gain for the controller, with the gain increasing inversely as the rolling period decreases. The key advantage of using the control law in Eq.~\eqref{eq:control} is its simplicity in implementation within Simulink and ease of tuning, even compared to the PID controller.
%%%%%%%%%%%%%%%%%%%%%%%%%%%%%%%%%%%%%%%%%%%%%%%%%%%%%%%%%%%%%%%%%%%%%%%%%%%%%%%%%%%%%%%%%%%%%%%%%%%%%%%%%%%
\subsection{Trajectory}
The goal of this work is to study robot-to-robot handover onboard a free-floating SMS. To accomplish this, a handover scenario is created where manipulator $A$ grabs a floating CubeSat (3U, zero initial velocity) in space and hands it to manipulator $B$. Both manipulators are given a schedule of end-effector poses, with the assumption that the CubeSat will be grasped correctly (the task schedule for the handover is described later in this work). The pose schedule is designed to demonstrate the handover process and is not guaranteed to be optimal.

IK is solved throughout the simulation to obtain up-to-date manipulator configurations for the desired end-effector pose. These calculations are necessary due to the base spacecraft's reaction to manipulator motion. MATLAB's \emph{Robotic System Toolbox} provides built-in IK solvers that run adequately fast during simulation. Among the solver options offered, this work uses the Levenburg-Marquardt (LM) method, a least-squares approach that incorporates a damping factor to choose the weighting factors of the constraints while solving the IK equations \cite{SUGIHARA2011}. The joints of the base spacecraft must be locked during the IK calculation to ensure that the solver does not return a configuration the SMS cannot physically reach. The complete configuration information (i.e., all $6+2n$ joint states) is also required as the initial guess for the LM solver to find a solution close to the current state. This approach increases the likelihood that the desired, online-generated trajectory avoids issues associated with redundancy, though such avoidance is not guaranteed.

This trajectory generation method assigns fixed time intervals for task execution, in contrast to approaches that optimize for time minimization while assigning tasks. Future work aims to develop more robust trajectory generation techniques that account for task retries, trajectory optimization, and obstacle avoidance.

%%%%%%%%%%%%%%%%%%%%%%%%%%%%%%%%%%%%%%%%%%%%%%%%%%%%%%%%%%%%%%%%%%%%%%%%%%%%%%%%%%%%%%%%%%%%%%%%%%%%%%%%%%%
\section{Simulation Study}
\begin{figure}[!t]
    \centering
    \includegraphics[width=0.77\linewidth]{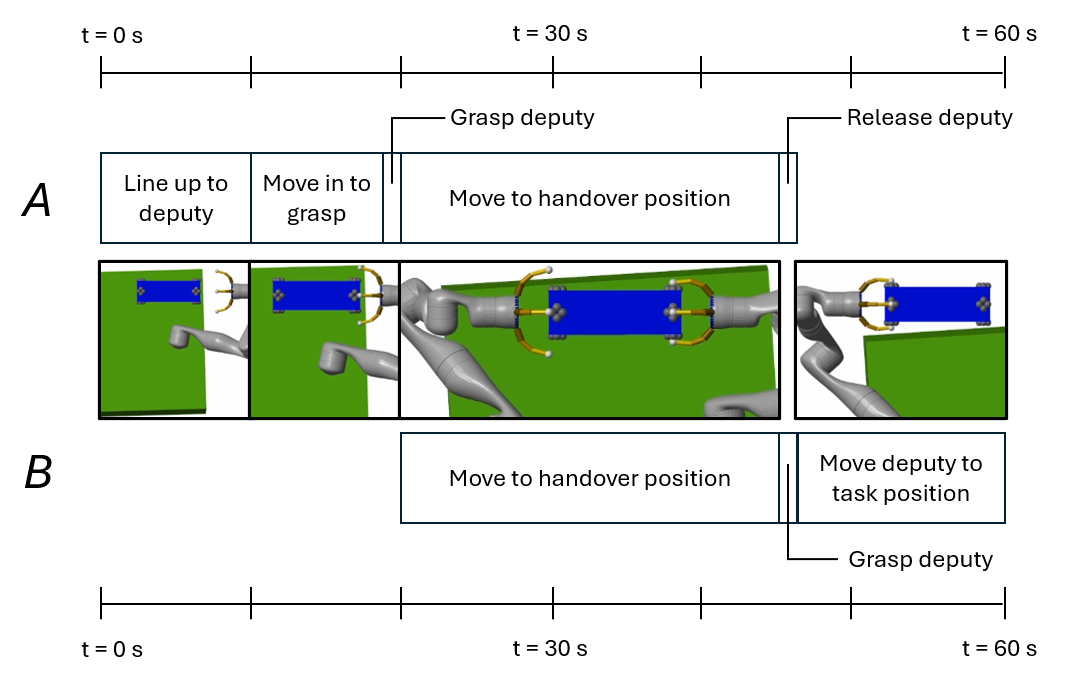}
    \caption{Planned Sequence of the Handover Operation between Dual Manipulators}
    \label{fig:schedule}
\end{figure}

A complete task schedule for both manipulators is shown in Figure \ref{fig:schedule}. Only manipulator $A$ is active during the first third of the operation as it moves in to grasp the deputy spacecraft. Meanwhile, manipulator $B$ is commanded to maintain its initial configuration while the base spacecraft reacts to the motion of manipulator $A$. After manipulator $A$ grasps the deputy, both manipulators move to handover location. Simultaneously, manipulator $A$ releases the deputy, while manipulator $B$ grasps it, thereby avoiding the formation of a closed kinematic chain. Manipulator $B$ then moves to an arbitrary pose to demonstrate that it has gained control of the deputy. The manipulators are given extra time for each tasks to ensure they can achieve the desired pose before proceeding. All poses were chosen offline for the purpose of demonstrating the handover process.

Simulation parameters are summarized in Table \ref{tab:params}. Due to the low manipulators-to-base mass ratio, approximately $6~\%$, the dynamic coupling effect is limited. The rolling period, and thus the NMPC control gain, is adjusted from $1.0$ to $2.0$ once the deputy is grasped to keep control effort low. High control effort significantly slows down the simulation, so scheduling the gain values helps manage simulation time. The PID gains were determined experimentally to ensure task completion within the allotted time frames.

\begin{table}[!t]
	\fontsize{10}{10}\selectfont
    \caption{Simulation Parameters}
   \label{tab:params}
        \centering 
   \begin{tabular}{ c | c | c | c } % Column formatting, 
      \hline 
      \textbf{Parameter}              & \textbf{Variable} & \textbf{Value}    & \textbf{Unit}\\
      \hline 
      Base Mass                       & $m_b$             & 240               & kg \\
      \hline
      Arm Mass                        & $m_m$             & 7                 & kg \\
      \hline
      Simulation Time                 & $T_s$             & 60                & s  \\
      \hline
      Rolling Period                  & $T_r$             & 1.0, 2.0          & s  \\
      \hline
      Proportional Gain               & $K_P$             & 2.0               & -  \\
      \hline
      Integral Gain                   & $K_I$             & 1.0               & -  \\
      \hline
      Derivative Gain                 & $K_D$             & 1.5               & -  \\
      \hline
      Absolute and Relative Tolerance & -                 & $1\times10^{-4}$  & -  \\
      \hline
   \end{tabular}
\end{table}

End-effector position error time histories for both manipulators are shown in Figures 4 and 5, with annotations of the handover operation. Sudden increases in error correspond to change in goal positions, as defined in the task schedule. For example, at $t=20$ s, manipulator $A$, having just grasped the deputy, is commanded to move to the handover location. 

The simulation was conducted using both the NMPC and PID controllers. Both controllers successfully completed the mission objective{, despite PID being theoretically unsuited for the SMS}. However, the PID controller, using non-optimized gain, exhibited greater overshoot and longer settling times than NMPC.

Figures \ref{fig:base}-\ref{fig:qb} show the joint state time histories of the base spacecraft, Arm $A$, and Arm $B$, respectively. In the case of the PID controller, the base spacecraft experiences drift after the end-effectors reach their goal poses. To compensate for this motion, the manipulators also drift, updating their configurations based on the IK solver. This issue is not present with the NMPC controller. Figures \ref{fig:ua} and \ref{fig:ub} show the control torque for Arm $A$ and Arm $B$, respectively. The maximum joint torque used by either controller is around $2.0$ Nm. Both control laws exhibited similar effort despite significantly diverging end-effector trajectories. This is possible due to strong controller impulses followed by minimal effort.

\begin{figure}[!t]
    \centering
    \includegraphics[width=0.9\linewidth]{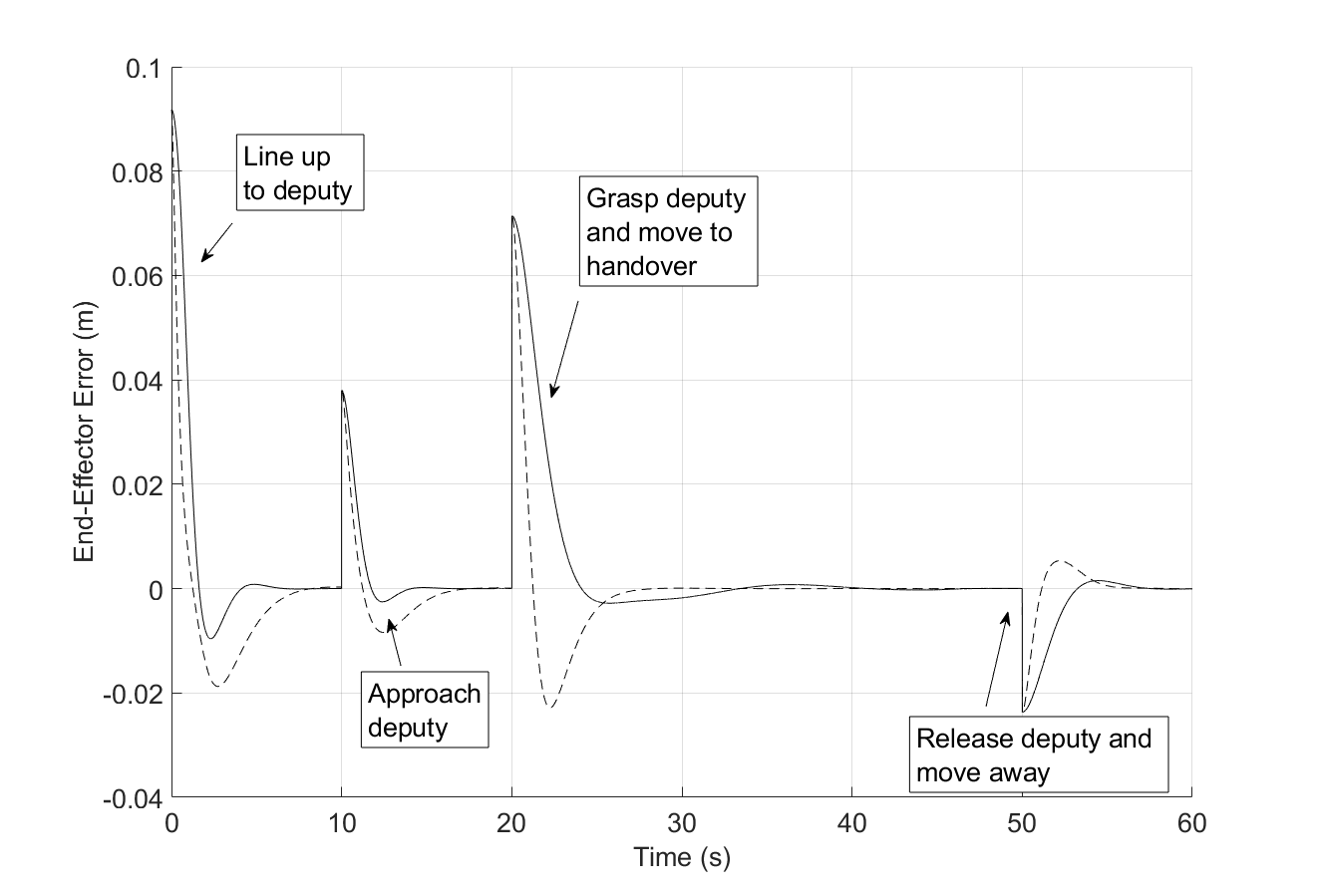}
    \caption{Position Error of the End-Effector $A$ (Solid Line: NMPC, Dashed Line: PID)}
    \label{fig:eea}
\end{figure}

\begin{figure}[!t]
    \centering
    \includegraphics[width=0.9\linewidth]{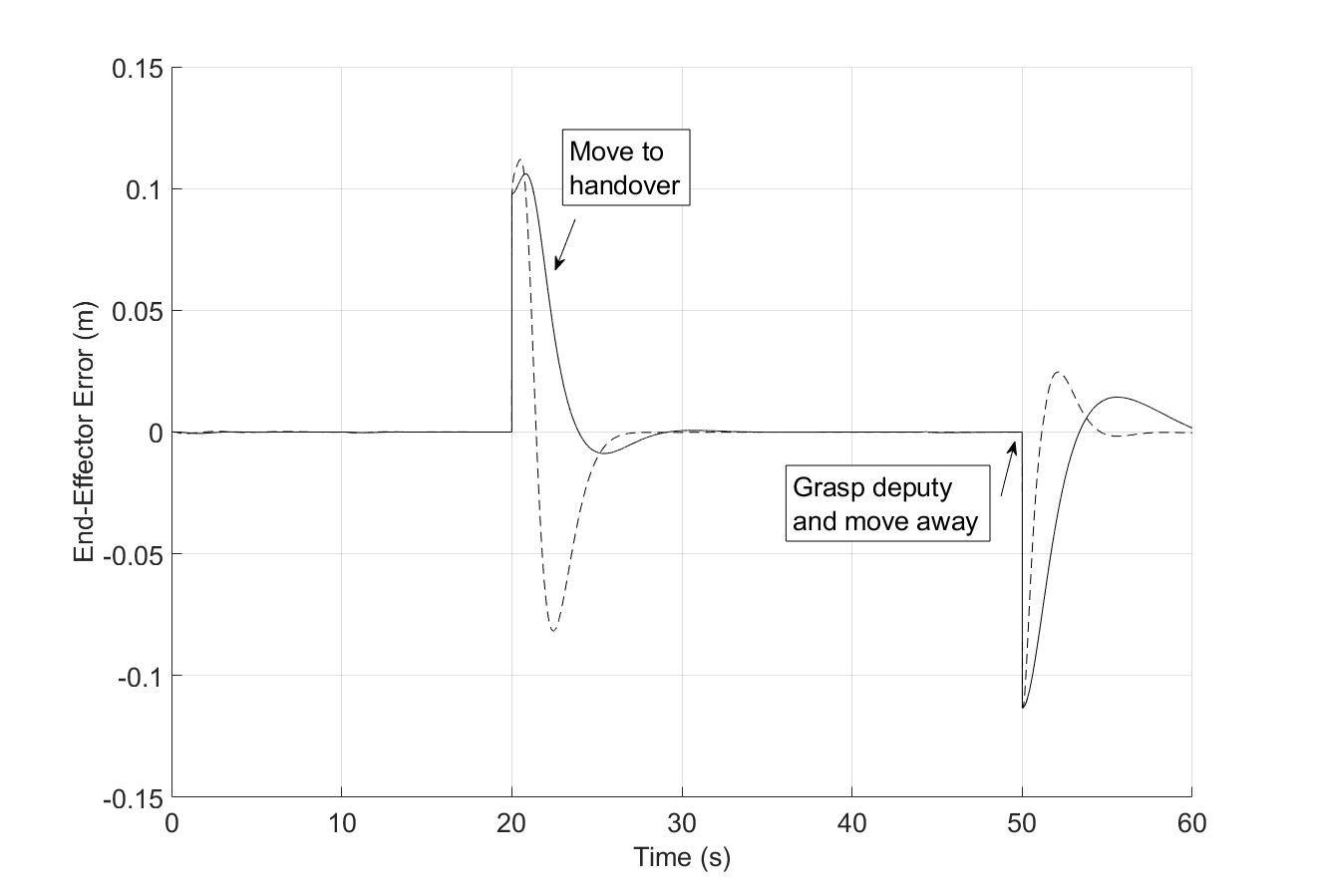}
    \caption{Position Error of the End-Effector $B$ (Solid Line: NMPC, Dashed Line: PID)}
    \label{fig:eeb}
\end{figure}

\begin{figure}[!t]
    \centering
    \includegraphics[width=0.9\linewidth]{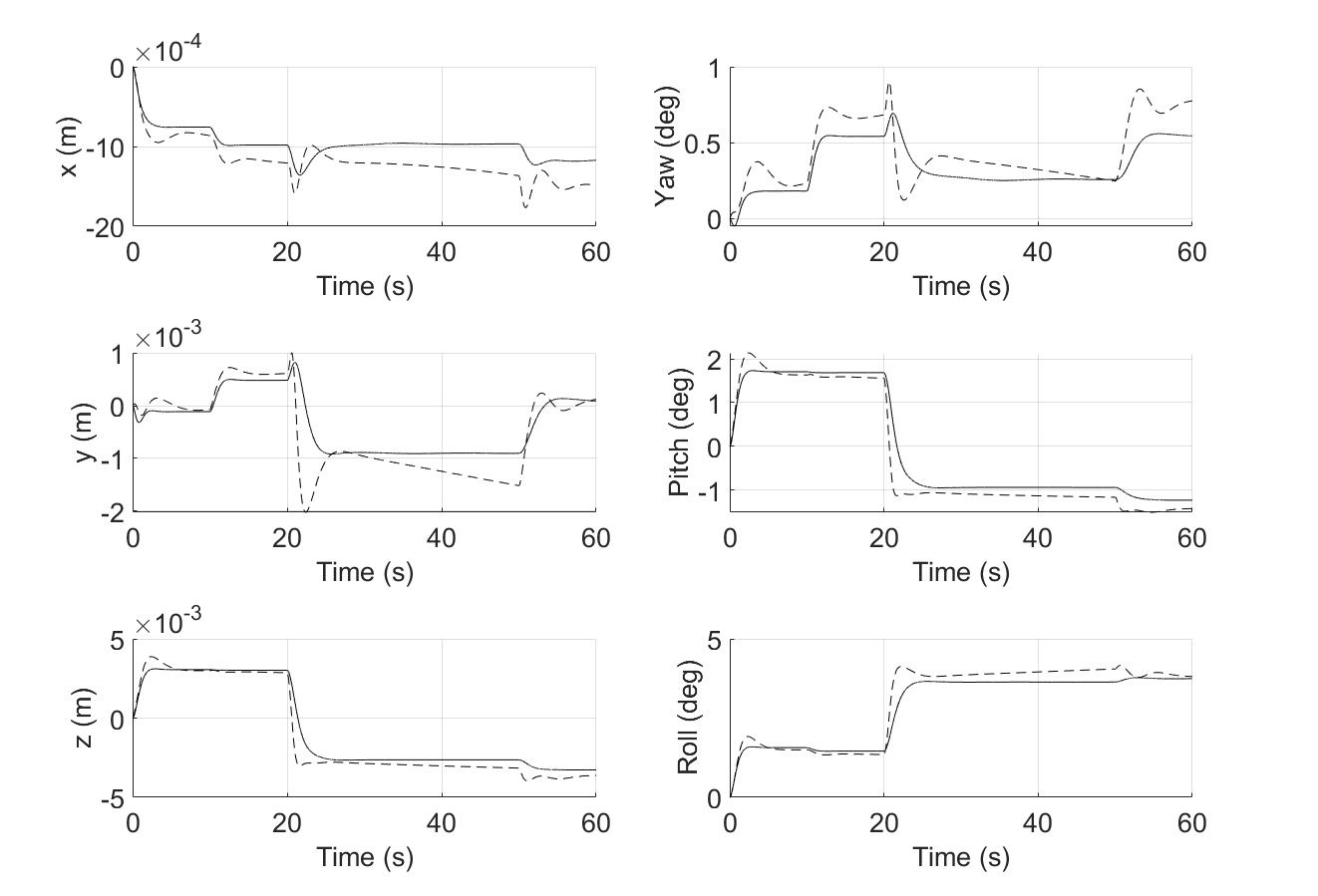}
    \caption{Position and Attitude of the Base Spacecraft (Solid Line: NMPC, Dashed Line: PID)}
    \label{fig:base}
\end{figure}

\begin{figure}[!t]
    \centering
    \includegraphics[width=0.9\linewidth]{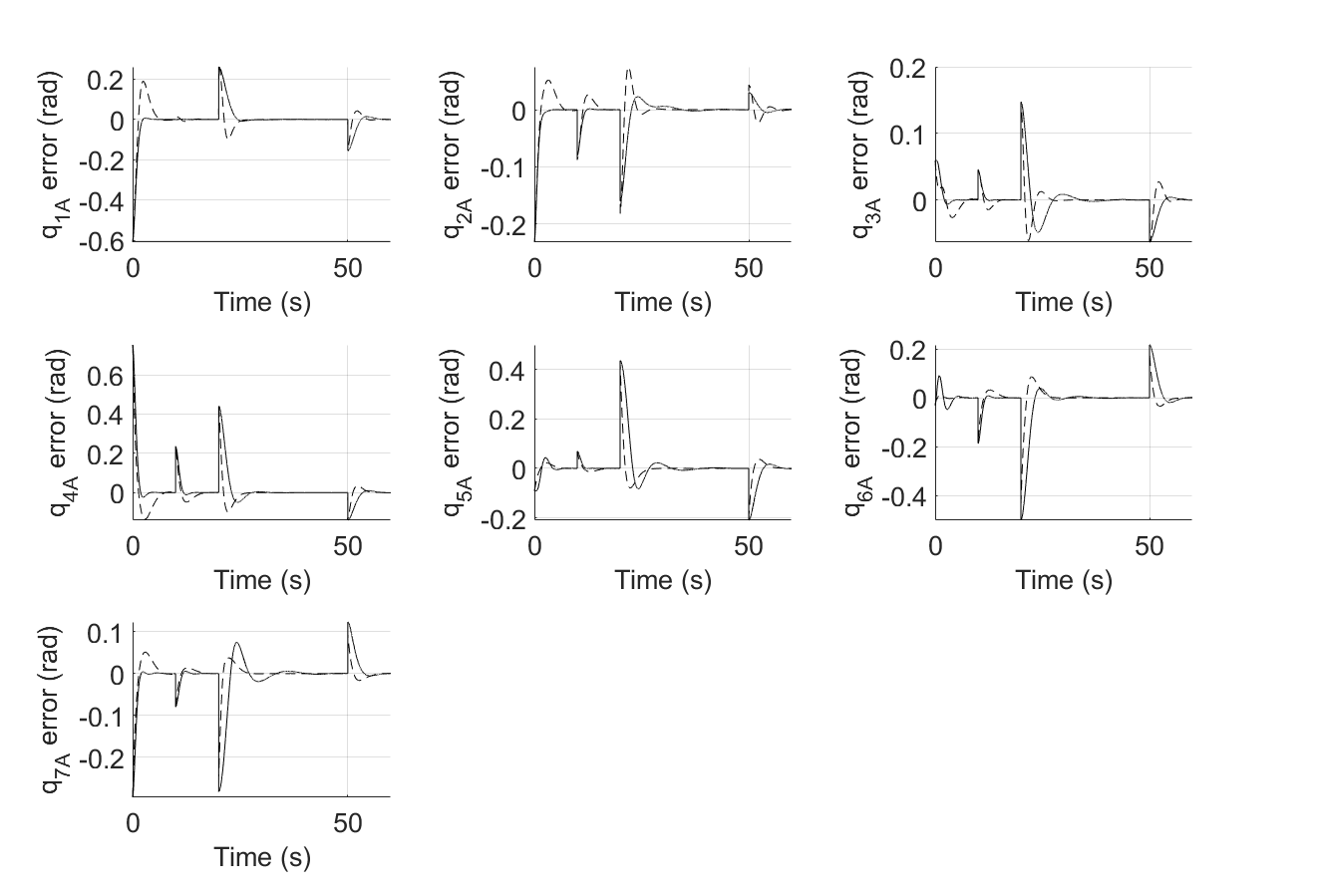}
    \caption{Joint Error of the Manipulator $A$ (Solid Line: NMPC, Dashed Line: PID)}
    \label{fig:qa}
\end{figure}

\begin{figure}[!t]
    \centering
    \includegraphics[width=0.9\linewidth]{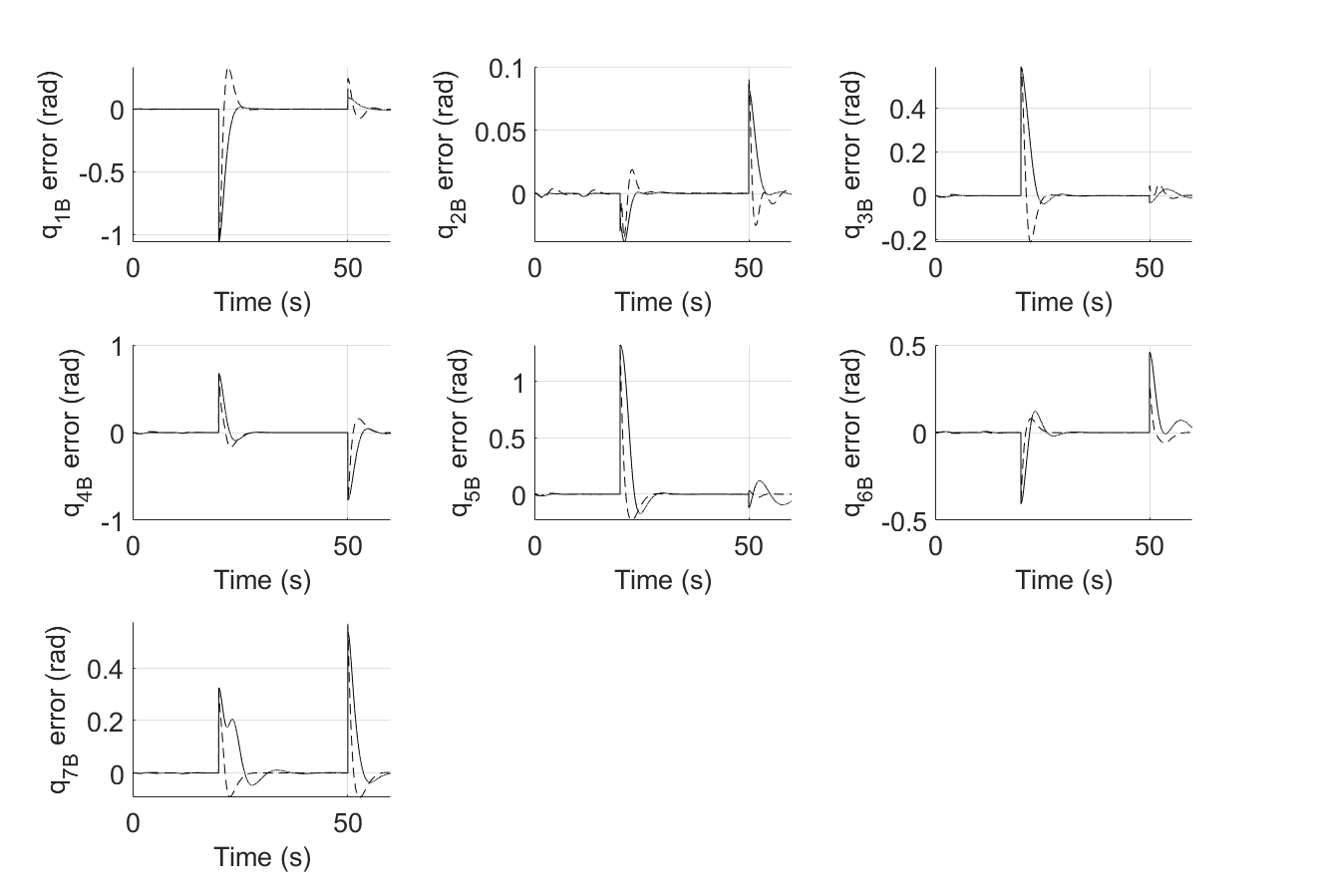}
    \caption{Joint Error of the Manipulator $B$ (Solid Line: NMPC, Dashed Line: PID)}
    \label{fig:qb}
\end{figure}

\begin{figure}[!t]
    \centering
    \includegraphics[width=0.9\linewidth]{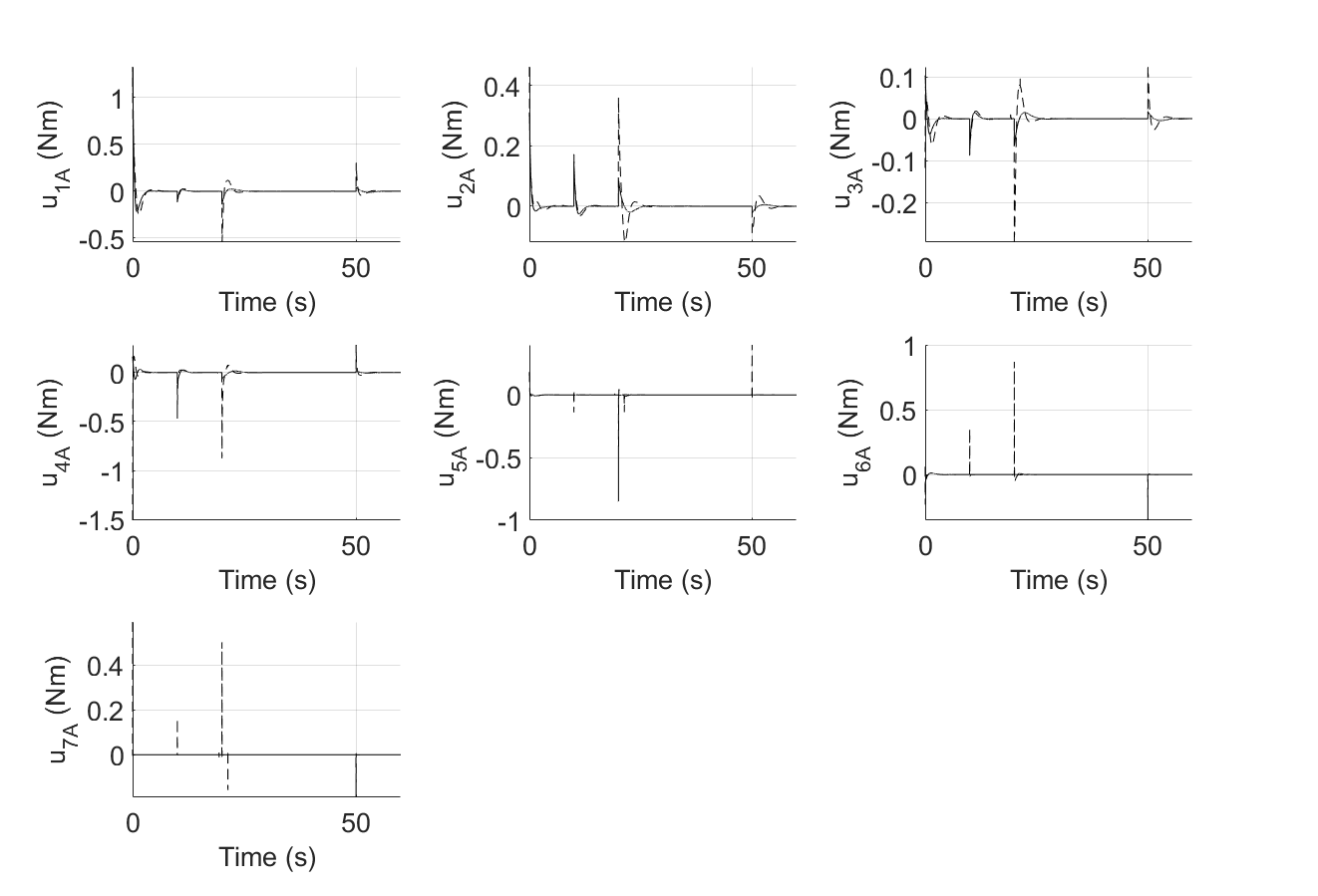}
    \caption{Control Effort of the Manipulator $A$ (Solid Line: NMPC, Dashed Line: PID)}
    \label{fig:ua}
\end{figure}

\begin{figure}[!t]
    \centering
    \includegraphics[width=0.9\linewidth]{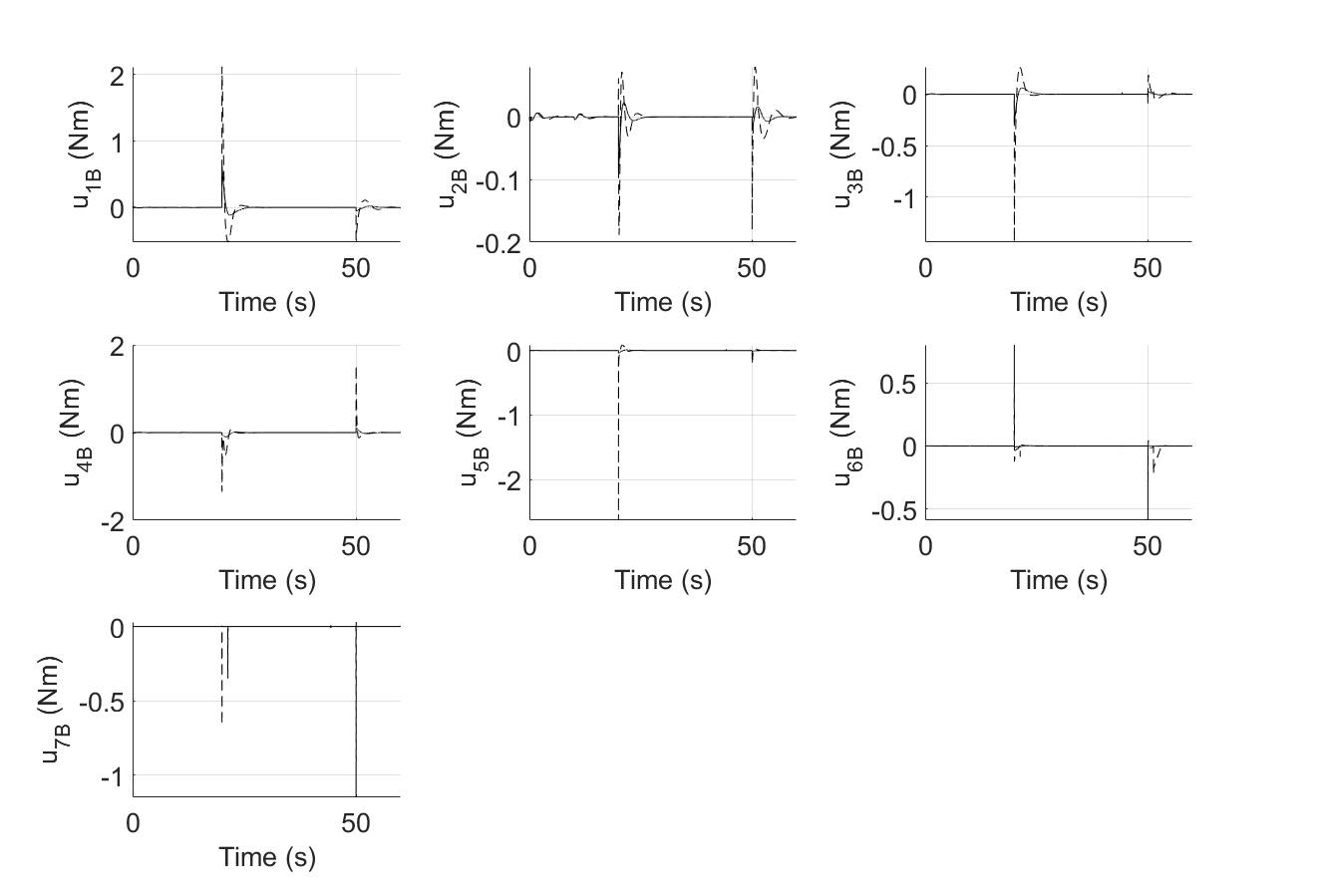}
    \caption{Control Effort of the Manipulator $B$ (Solid Line: NMPC, Dashed Line: PID)}
    \label{fig:ub}
\end{figure}

%%%%%%%%%%%%%%%%%%%%%%%%%%%%%%%%%%%%%%%%%%%%%%%%%%%%%%%%%%%%%%%%%%%%%%%%%%%%%%%%%%%%%%%%%%%%%%%%%%%%%%%%%%%
\clearpage
\section{Conclusion}
This work demonstrates the modeling and control of a high degree-of-freedom space manipulator system capable of executing accurate trajectory tracking through both nonlinear model predictive and proportional-integral-derivative control schemes. The simulation results show that the controllers can effectively manage end-effector motion as the base spacecraft reacts dynamically. Future work will focus on extending this study by testing the controllers in more complex rendezvous-to-rendezvous handover scenarios, considering uncertainties such as disturbances, model inaccuracies, and sensor noise. This will allow for a more comprehensive evaluation of the controllers’ robustness, adaptability, and performance in dynamic, real-world conditions, which are critical for the successful implementation of in-space servicer systems.
%%%%%%%%%%%%%%%%%%%%%%%%%%%%%%%%%%%%%%%%%%%%%%%%%%%%%%%%%%%%%%%%%%%%%%%%%%%%%%%%%%%%%%%%%%%%%%%%%%%%%%%%%%%
\section{Acknowledgment}

This material is based on research sponsored by Air Force Research Laboratory (AFRL) under agreement number FA9453-24-2-0004. The U.S. Government is authorized to reproduce and distribute reprints for Governmental purposes notwithstanding any copyright notation thereon. The views and conclusions contained herein are those of the authors and should not be interpreted as necessarily representing the official policies or endorsements, either expressed or implied, of Air Force Research Laboratory (AFRL) and or the U.S. Government.
%%%%%%%%%%%%%%%%%%%%%%%%%%%%%%%%%%%%%%%%%%%%%%%%%%%%%%%%%%%%%%%%%%%%%%%%%%%%%%%%%%%%%%%%%%%%%%%%%%%%%%%%%%%
\bibliographystyle{AAS_publication}   % Number the references.
\bibliography{references}   % Use references.bib to resolve the labels.
\end{document}